\documentclass[lettersize,journal]{IEEEtran}
\usepackage{amsmath,amsfonts}
\usepackage{algorithmic}
\usepackage{algorithm}
\usepackage{array}
\usepackage[caption=false,font=normalsize,labelfont=sf,textfont=sf]{subfig}
\usepackage{textcomp}
\usepackage{stfloats}
\usepackage{url}
\usepackage{verbatim}
\usepackage{graphicx}
\usepackage{cite}
\usepackage{bbm}
\usepackage{multirow}
\usepackage{booktabs}
\usepackage{hyperref}
\usepackage{threeparttable}
\begin{document}

\title{Generalizing Segmentation Foundation Model Under Sim-to-real Domain-shift for Guidewire Segmentation in X-ray Fluoroscopy}

\author{Yuxuan Wen, Evgenia Roussinova, Olivier Brina, Paolo Machi, and Mohamed Bouri, \IEEEmembership{Senior Member, IEEE}
\thanks{Yuxuan Wen is with the Rehassist Group, EPFL, Lausanne, Switzerland, Department of Electrical Engineering and Computer Science, University of Cincinnati, OH 45221, USA, and also with the College of Computer Science, Chongqing University, Chongqing 400044, China (e-mail: wenyu@mail.uc.edu).}
\thanks{Evgenia Roussinova is with the Translational Neural Engineering Laboratory (TNE), EPFL, Lausanne, Switzerland (e-mail: evgenia.roussinova@epfl.ch).}
\thanks{Olivier Brina is with the Department of Interventional Neuroradiology, Geneva University Hospital, Geneva, Switzerland (e-mail: Olivier.Brina@hug.ch).}
\thanks{Paolo Machi is with the Department of Interventional Neuroradiology, Geneva University Hospital, Geneva, Switzerland, and also with the Brain Endovascular Therapeutics R\&D Lab, Campus Biotech, Geneva, Switzerland (e-mail: Paolo.Machi@unige.ch, Paolo.Machi@hug.ch).}
\thanks{Mohamed Bouri is with the Translational Neural Engineering Laboratory (TNE) and the Biorobotics Laboratory (Biorob), EPFL, Lausanne, Switzerland (e-mail: mohamed.bouri@epfl.ch).}}
\markboth{arxiv preprint}%
{Y. Wen \MakeLowercase{\textit{et al.}}: Generalizing Segmentation Foundation Model Under Sim-to-real Domain-shift for Guidewire Segmentation in X-ray Fluoroscopy}


\maketitle

\begin{abstract}

Guidewire segmentation during endovascular interventions holds the potential to significantly enhance procedural accuracy, improving visualization and providing critical feedback that can support both physicians and robotic systems in navigating complex vascular pathways. Unlike supervised segmentation networks, which need many expensive expert-annotated labels, sim-to-real domain adaptation approaches utilize synthetic data from simulations, offering a cost-effective solution. The success of models like Segment-Anything (SAM) has driven advancements in image segmentation foundation models with strong zero/few-shot generalization through prompt engineering. However, they struggle with medical images like X-ray fluoroscopy and the domain-shifts of the data. Given the challenges of acquiring annotation and the accessibility of labeled simulation data, we propose a sim-to-real domain adaption framework with a coarse-to-fine strategy to adapt SAM to X-ray fluoroscopy guidewire segmentation without any annotation on the target domain. We first generate the pseudo-labels by utilizing a simple source image style transfer technique that preserves the guidewire structure. Then, we develop a weakly supervised self-training architecture to fine-tune an end-to-end student SAM with the coarse labels by imposing consistency regularization and supervision from the teacher SAM network. We validate the effectiveness of the proposed method on a publicly available Cardiac dataset and an in-house Neurovascular dataset, where our method surpasses both pre-trained SAM and many state-of-the-art domain adaptation techniques by a large margin. Our code will be made public on \href{https://github.com/Yuxuan-Wen/Sim2real-Guidewire-Seg}{GitHub} soon.

\end{abstract}


\section{Introduction}

\begin{figure}[t]
  \centering
  \includegraphics[width= \linewidth]{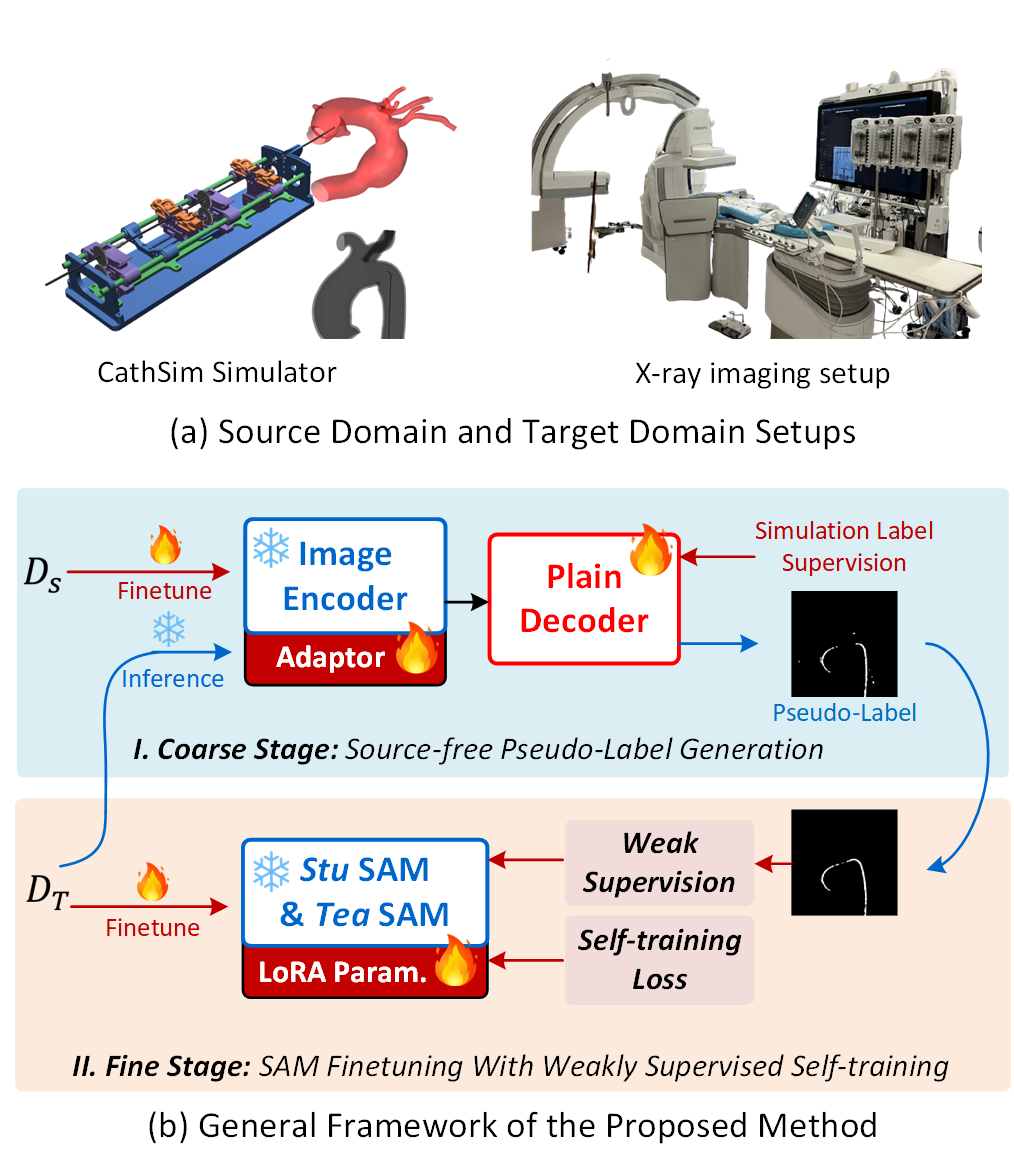}
  \caption{Problem formulation and method overview. (a) The setups in the source domain and target domain. (b) A general framework of the proposed coarse-to-fine sim-to-real domain adaptation method, where we generate the pseudo-labels in the coarse stage and utilize weak supervision and self-training to train a student-teacher SAM network. The ``fire'' symbol means the trainable parameters, while the  ``snowflake'' symbol means the parameters are frozen without updating.}
  \label{fig1}
\end{figure}

\IEEEPARstart{E}{ndovascular} interventions \cite{derex2017mechanical, yoo2017thrombectomy} are minimally invasive procedures conducted within blood vessels using a guidewire \cite{qiu2022guidewire}, an essential tool that is inserted into the cardiovascular system through per-cutaneous vascular access and is used to navigate the catheter through the vessels under real-time X-ray fluoroscopy guidance with the final aim of putting the catheter tip at the desired location (\textit{i.e.} Aneurysm ostium or through the thrombus). Precise navigation of the guidewire is crucial to avoid damaging vascular structures \cite{rafii2014current, simaan2018medical} and prevent complications. Accurate guidewire segmentation in consecutive X-ray fluoroscopy images can significantly aid the navigation process by reducing the interventionist's workload \cite{mazomenos2016survey}, providing visual feedback to robotic systems \cite{liu2024image}, and even enabling semi- or fully automatic robot-assisted interventions. Although supervised deep learning segmentation methods have obtained impressive results\cite{roy2023mednext,guo2022unet,cciccek20163d}, a large scale of labeled data is required, which is extremely expensive and time-consuming to obtain. In contrast to fully supervised learning, sim-to-real domain adaption offers an alternative solution by collecting simulated data with annotation and generalizing the model to real-world data by bridging the domain gap. Though the cost and effort for data labeling is eliminated, such transfer presents challenges to the zero- or few-shot ability of the model. For example, in the case of the segment anything model (SAM) \cite{kirillov2023segment}, despite the vast size of the large-scale natural datasets used for its pre-training, SAM has been observed to perform awkwardly on certain out-of-distribution downstream tasks \cite{chen2023sam} due to the distribution shift between the training dataset and testing dataset. Prior research transfers SAM to new tasks by finetuning \cite{wu2304medical, qiu2023learnable, hu2021lora}, which still requires annotation of the target domain. There exist methods that explore SAM with semi-supervised learning \cite{miao2024cross} and self-training with an anchor model \cite{zhang2024improving} to minimize the annotation cost, but SAM models that require prompts to generate prediction results fail to meet the real-time application requirements set by endovascular interventions. In addition, guidewire segmentation is a distinct task that stands apart from other medical image segmentation tasks due to its unique structure, solely relying on prompted results as supervision would bring unsatisfying results due to the long-thin shape of the guidewire.

To tackle the above limitations, we propose a sim-to-real domain adaptation framework for guidewire segmentation that enables the usage of simulation data to generalize an end-to-end segmentation foundation model to real-world X-ray fluoroscopy without the need for any annotations. A problem formulation illustration of the setup in both the target domain (real-world) and source domain (simulation) and a general diagram of the proposed method can be found in Fig. \ref{fig1}. Our framework follows a coarse-to-fine fashion consisting of two stages to exploit weak supervision and unlabeled data on the target domain. We use the synthesized images generated by the CathSim simulator \cite{jianu2024cathsim} as the source domain $D_S$, and the images captured by a real-world X-ray imaging system as the target domain $D_T$. To be more specific, we first rely on the generalizability of the pre-trained SAM and employ source-free domain adaption to generate the pseudo labels by quickly finetuning an image encoder and a plain decoder with synthesized target-like data. In the following fine stage, we use the coarse labels as weak supervision and develop a self-training architecture to fine-tune a prompt-free end-to-end student SAM model by imposing both embedding-level and prediction-level consistency regularizations and the guidance of an independent teacher SAM model that takes in prompts to produce superior results than the weak supervision. During the self-training process, the guided teacher model converges first and then immediately supervises the student model to achieve the same level of performance.

We validate the effectiveness of the proposed method using a publicly available Cardiac dataset \cite{gherardini2020catheter}, and an in-house X-ray fluoroscopic Neurovascular dataset collected in Geneva University Hospitals (HUG). Results show that we significantly outperform both pre-trained SAM and many state-of-the-art domain adaptation techniques by a distinguishable margin. In summary, this paper offers the following contributions:

\begin{itemize}

\item A sim-to-real coarse-to-fine domain adaption framework is proposed to generalize a prompt-free end-to-end vision foundation model to real-world X-ray fluoroscopy without the need for any forms of annotation.

\item We propose a pipeline that generates coarse pseudo-labels through SAM model finetuning by utilizing source image style transfer that preserves the guidewire structure.

\item We exploit weak supervision from pseudo-labels and self-training losses to further finetune a teacher-student SAM framework to achieve superior results than the coarse labels.

\item Extensive experiment on two X-ray fluoroscopic phantom datasets demonstrates the effectiveness of the proposed weakly supervised self-training adaptation approach.

\end{itemize}

\section{Related Works}

\subsection{Guidewire Segmentation}

Guidewire segmentation has advanced significantly with the advent of deep learning frameworks \cite{ambrosini2017fully, zhou2020real}. The FW-Net \cite{nguyen2020end} employed an end-to-end approach with an encoder-decoder structure, optical flow extraction, and a unique flow-guided warping function to maintain temporal continuity in imaging sequences. Gherardini \textit{et al.} \cite{gherardini2020catheter} proposed a transformative approach that integrates Convolutional Neural Networks (CNNs) with transfer learning, utilizing synthetic fluoroscopic images to train a streamlined segmentation model that requires minimal manual annotations. In addition to these methods, \cite{wu2018automatic, li2019two, zhang2022real} employed a two-stage framework where a detector is first used to output the bounding box of the guidewire, which helps to reduce class imbalance in the segmentation task. In contrast to advancements in previous research, which all require labeled data for supervision, we perform guidewire segmentation in a source-free sim-to-real domain adaptation setting. Our framework eliminates the need for annotation labels while simultaneously ensuring performance. To the best of our knowledge, we are the first to explore sim-to-real domain adaptation for guidewire segmentation tasks.

\subsection{Unsupervised Domain Adaptation}

Early Unsupervised Domain Adaptation (UDA) methods concentrate on aligning feature distributions between source and target domains, whereas deep UDA methods prioritize learning domain-invariant features \cite{zhao2020review}.  Within this category, discrepancy-based methods \cite{chen2020homm} incorporated a loss function designed to minimize the discrepancy between the prediction streams, while adversarial discriminative methods \cite{wulfmeier2017addressing} achieved the same goal with the adversarial objective. In addition, adversarial generative methods \cite{zhu2017unpaired} integrated generative adversarial networks (GANs) with discriminators at either the image or feature level. Many works have explored image-to-image translation \cite{liu2017unsupervised, huang2018multimodal} or style transfer \cite{gong2022one, luo2020adversarial} to learn a shared latent space between domains or close the domain gap in the image level. TENT \cite{wang2020tent} and TRIBE \cite{su2024towards} adapt a source domain model to testing data during inference by means of Test-Time Adaptation (TTA), demonstrating the success with unseen corruptions. In this paper, we use a straightforward source image style transfer method that maintains the guidewire structure and performs UDA with the few-shot capabilities of the SAM model.

\subsection{Image Segmentation Foundation Model}

In recent years, foundation models have revolutionized artificial intelligence by leveraging web-scale datasets \cite{brown2020language, radford2021learning} for pre-training to achieve impressive zero-shot generalization across various scenarios. This motivates the emergence of vision foundation models like SAM \cite{kirillov2023segment} and DINO v2 \cite{oquab2023dinov2}, which are pre-trained on huge datasets. Among these models, SAM stands out for its ability to enable zero-shot segmentation using prompt engineering. Many attempts have been made to validate the robustness of SAM under more challenging scenarios in medical images \cite{ma2024segment, zhang2023customized, wu2304medical}. Recently, many works have adopted SAM in settings with limited annotations. CPC-SAM \cite{miao2024cross} introduced a cross-prompting consistency method that leverages SAM within a dual-branch framework, enabling effective semi-supervised medical image segmentation. WeSAM \cite{zhang2024improving} proposed a self-training strategy with an anchor model that adapts SAM to various target distributions, improving its robustness and efficiency across multiple segmentation tasks. Despite these initial efforts to identify weaknesses, there are very few mature solutions available to enhance SAM's generalization from simulation data to real-world data.

\begin{figure*}[t]
  \centering
  \includegraphics[width= \linewidth]{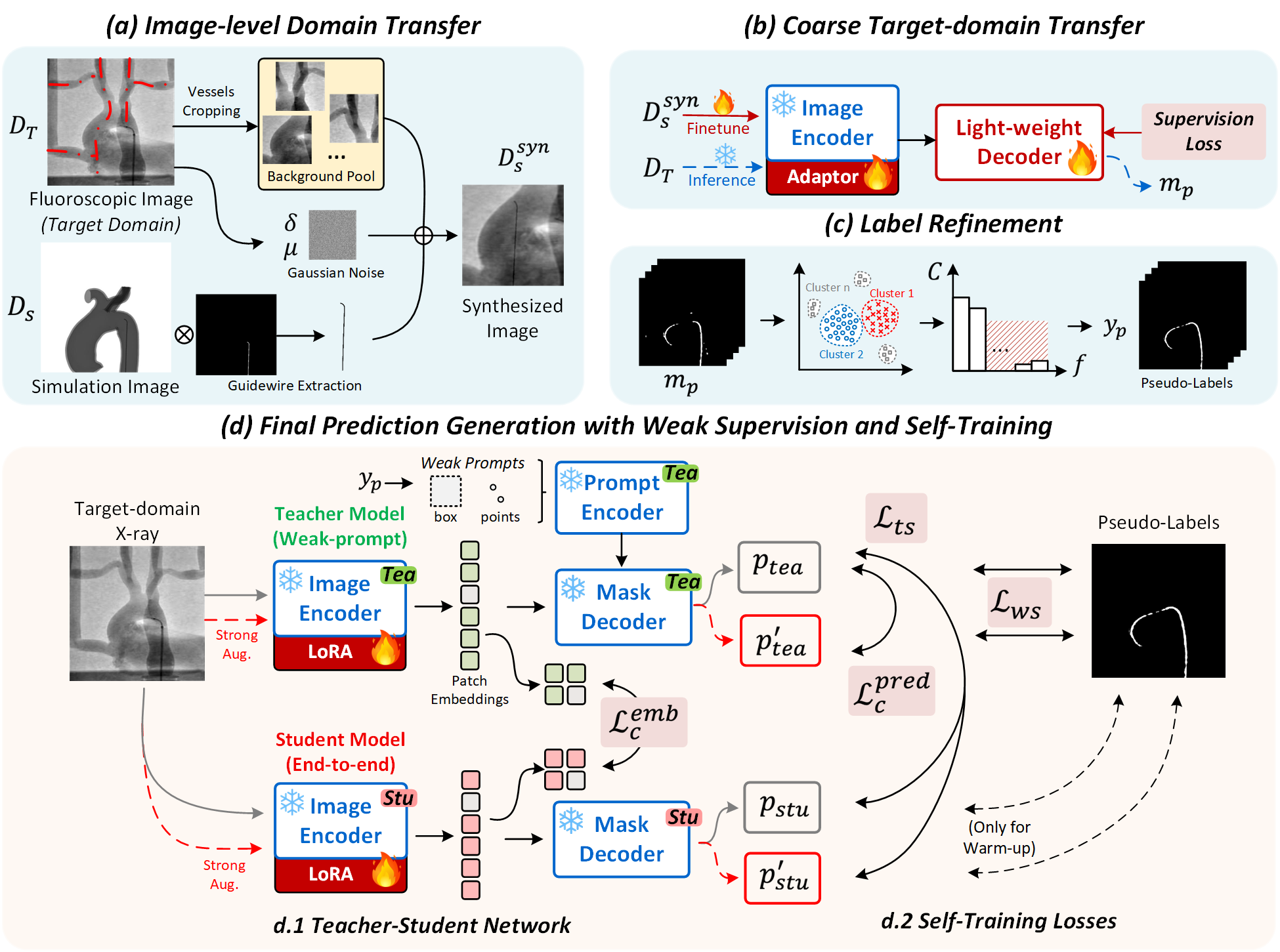}
  \caption{The proposed sim-to-real adaptation framework, which contains a coarse stage ((a)-(c)) that prepares the pseudo-labels for weak supervision and a fine stage (d) that generates the final prediction by finetuning a student-teacher network with self-training losses. ``Weak prompts'' are generated from pseudo-labels in the form of either point or box during training. The two types of prompts, as well as their combination, were evaluated (see Section \ref{IV.F.3}). ``Tea'' means the teacher SAM and ``Stu'' means the student SAM. The ``fire'' symbol means the trainable parameters, while the  ``snowflake'' symbol means the parameters are frozen without updating.}
  \label{fig_overall}
\end{figure*}

\section{Methodology}

For our methodology, we first discuss in Section \ref{sec1} the preliminaries of the Segment Anything Model \cite{kirillov2023segment}, and provide an overview of the problem formulation and our overall pipeline in Section \ref{sec2}. Then, we describe our motivation and detailed implementation of the image-level transfer and pseudo-label generation in Section \ref{sec3}, student-teacher network in Section \ref{sec4}, and the weakly-supervised self-training in \ref{sec5}.

\subsection{Preliminaries: Segment Anything Model}
\label{sec1}

We utilize the segment-anything model \cite{kirillov2023segment} as the skeleton of our framework. SAM consists of three major components, 1) the image encoder $z = f(x; \mathit{\Delta})$ pre-trained using a masked autoencoder (MAE) \cite{he2022masked} to extract feature embeddings from the image, 2) prompt encoder $e = g(w; \mathit{\Omega})$ to output prompt embeddings for given prompt, and 3) mask decoder $h(z, e; \mathit{\Phi})$ to produce segmentation results based on the feature embeddings $z$ and prompt embeddings $e$. Here, $\mathit{\Delta}$, $\mathit{\Omega}$ and $\mathit{\Phi}$ are weights for each of the components. The SAM model is further fine-tuned on the SA-1B dataset \cite{kirkpatrick2017overcoming}, a web-scale labeled training set consisting of 1.1 billion labeled masks. In our work, we employ a Learnable Ophthalmology SAM \cite{qiu2023learnable} in the coarse stage for its fast-convergence ability and a teacher-student architecture composed of two LoRA models \cite{hu2021lora} in the fine stage, since we want to utilize prompts as guidance for the teacher model and this model has been proven effective by several works \cite{zhang2024improving, miao2024cross}. By utilizing weak supervision from the coarse pseudo-labels and self-training losses, our approach generalizes SAM for guidewire segmentation in X-ray fluoroscopy, effectively handling sim-to-real domain shifts.

\subsection{Problem Formulation and Overall Framework}
\label{sec2}

Sim-to-real domain adaption aims to obtain a satisfactory performance on the unlabeled real-world target domain \( D_{T}=\left\{\left(x_{i}^{T}\right)\right\}_{i=1}^{N_T} \) using labeled simulation source domain \( D_{S}=\left\{\left(x_{i}^{S}, y_{i}^{S}\right)\right\}_{i=1}^{N_S} \). \( x_{i} \in \mathcal{R}^{H \times W} \) is the 2D X-ray fluoroscopic images, and \( y_{i} \in \{0, 1\}^{H \times W} \) is the corresponding annotation for labeled data with 1 being the guidewire and 0 being the background.

Fig. \ref{fig_overall} provides an overview of our proposed two-stage weakly supervised self-training framework with SAM for sim-to-real domain adaptation. To make the most of the few-shot learning capabilities of the SAM model, we conduct an image-level transfer and directly generalize the finetuned SAM to acquire the pseudo-labels. Building on this foundation, a teacher-student network is developed based on the promptable property of SAM to fully utilize the weak supervision of the pseudo-labels. The student network benefits from the guidance of the teacher model and additional consistency constraints, resulting in improved outcomes compared to the coarse stage.

\subsection{Image-level Transfer and Pseudo-label Generation}
\label{sec3}

Source image style transfer is widely adapted as a means of bringing domain gap to depict the style of the target domain, most unsupervised domain adaption \cite{rafi2024domain} methods utilize stylization \cite{gong2022one, luo2020adversarial} or generate new target-alike images with diffusion models \cite{benigmim2023one}. However, these methods usually alter the content of the image, compromising the location of the guidewire. To preserve the guidewire's structure, we replicate the style of the target domain by directly cropping a target-domain background and incorporating the guidewire's location from the simulated source domain, along with domain-specific Gaussian noise. This process is shown in Fig. \ref{fig_overall} (a): we first construct a pool consisting of various background patches $B=\left\{\left(b_{k}\right)\right\}_{k=1}^{K}$, then extract the pixels of the guidewires from the simulation data by simple multiplication. For each frame, we impose the extracted guidewire pixels to a background randomly selected from the background pool: 

\begin{equation}
   x_{i}^{S^{\prime}} = (x_{i}^{S} \otimes y_{i}^{S}) \oplus b_{k}, b_{k} \in B
\end{equation}

In this way, we form a set of target-domain-alike synthesized images \( D_{S}^{syn}=\left\{\left(x_{i}^{S^{\prime}}, y_{i}^{S^{\prime}}\right)\right\}_{i=1}^{N} \), while still preserving the structure of the guidewire.

Considering the few-shot learning capabilities of the SAM model, we further finetune a SAM with a plain decoder for quick convergence in a source-free domain adaption manner. We first train the Learnable Ophthalmology SAM \cite{qiu2023learnable} with the synthesized dataset $D_{S}^{syn}$, and then perform inference on the target domain $D_{T}$ to generate $m_p$. The pseudo-labels $y_p$ are obtained by cleaning the raw inference output $m_p$ with DBSCAN clustering and cluster-wise filtering to reduce noise. This process, illustrated in Fig. \ref{fig_overall} (b) and (c), greatly increases the quality of the prompts for the teacher SAM model, which in return helps the convergence of the teacher model, simultaneously lifting the upper bound for the prompt-free student SAM model.

\subsection{Student-Teacher Network}
\label{sec4}

In the fine stage, illustrated in Fig. \ref{fig_overall} (d), we adopt a teacher-student architecture consisting of two models without any weight correlations: 1) a promptable teacher network $\mathcal{F}(x; \mathit{\theta^{t}})$ that takes in weak prompts as guidance, and 2) a prompt-free student network $\mathcal{F}(x; \mathit{\theta^{s}})$, which is trained as an end-to-end model, requiring no prompting to generate the segmentation results. With each input $x_{i} \in D_{T}$, we apply one strong data perturbation $\mathcal{P}$ to generate the augmented input $x_{i}^{\prime} = \mathcal{P}(x_{i})$ for both networks. 

Four prediction results $p_{stu}$, $p_{stu}^{\prime}$, $p_{tea}$, and $p_{tea}^{\prime}$ are obtained by passing $x$ and strongly augmented $x^{\prime}$ through teacher and student networks, respectively. For the student network, the prediction $p_{stu}$ is generated by incorporating both image embedding $z_{stu} = f(x; \mathit{\Delta^{stu}})$ based on input $x$ and prompt embedding $e_{stu} = g(t(y_p); \mathit{\Omega})$ based on weak prompts generated from the pseudo labels $y_p$: $p_{stu} = h(z_{stu}, e_{stu}; \mathit{\Phi})$. For the teacher network, $p_{tea}$ can be generated without any guidance of the prompt: $p_{tea} = h(z_{tea}; \mathit{\Phi})$, where $z_{tea} = f(x; \mathit{\Delta^{tea}})$. The predictions are further normalized using a sigmoid function and binarized with a threshold. Considering the difference in prompt method between the two models and the quick convergence of the teacher model, the weights of the tune-able parameters for student network $\mathit{\Delta^{stu}}$ and teacher network $\mathit{\Delta^{tea}}$ are not correlated in any way ($\mathit{\Delta^{stu}} \neq \mathit{\Delta^{tea}}$). This differs from many methods with the teacher-student architecture, where they usually adopt shared weights, using Exponential Moving Average (EMA) for the teacher model, or regulate the weights with Elastic Weight Consolidation (EWC) \cite{kirkpatrick2017overcoming}.

During training, we begin with a warm-up phase where supervision is applied using the pseudo-labels $y_p$, helping to initialize the student model and converge the teacher model. Afterward, the teacher model, with its parameters frozen, provides guidance to the student model.

\subsection{Weakly-supervised Self-training}
\label{sec5}

In order to achieve superior results than the coarse labels, we take advantage of the promotable property of the teacher SAM network by utilizing self-training losses. Self-training is a common technique in semi-supervised learning \cite{sohn2020fixmatch, zhao2023rcps, miao2024cross}, while also demonstrating its effectiveness in unsupervised domain adaptation task \cite{liang2021source, zhang2024improving}. In this work, we take advantage of the higher quality output from the teacher network, which is guided with prompts, to act as supervision to the student network using dice loss \cite{milletari2016v}. The teacher-student loss $\mathcal{L}_{ts}$ is calculated as follows:

\begin{equation}
\mathcal{L}_{\text {dice }}(x, \hat{y})=\sum_{i=1 \cdots N_T} 1-\frac{2 \sum_{h, w} x_{i h w} \cdot \hat{y}_{i h w}+\epsilon}{\sum_{h, w} x_{i h w}+\sum_{h, w} \hat{y}_{i h w}+\epsilon}
\end{equation}

\begin{align}
    \mathcal{L}_{ts} & = \lambda_{ts}\mathcal{L}_{dice}(p_{stu}, \mathit{Sig}(p_{tea})) + \lambda_{ts}^{\prime}\mathcal{L}_{dice}(p_{stu}^{\prime}, \mathit{Sig}(p_{tea}))
\end{align}

Where $\lambda$s are the hyper-parameters for balancing the loss, $\mathit{Sig}$ denotes the normalization with a sigmoid function and the binarization with a threshold, and $\epsilon$ is a very small constant preventing zero division.

Training networks solely with $\mathcal{L}_{ts}$ is prone to the accumulation of incorrect pseudo-labels predicted by the teacher network, leading to performance declines after extended iterations of self-training due to confirmation bias \cite{arazo2020pseudo, su2022revisiting}. Therefore, we incorporate weak supervision in addition to self-training by utilizing the pseudo-labels generated during the coarse stage. The weak supervision loss $\mathcal{L}_{ws}$ aims to effectively guide the learning progress, accelerate the convergence, and prevent the model from collapsing:

\begin{align}
    \mathcal{L}_{ws} &= \lambda_{ws}^{stu}\mathcal{L}_{dice}(p_{stu}, y_p) + \lambda_{ws}^{stu^{\prime}}\mathcal{L}_{dice}(p_{stu}^{\prime}, y_p) \nonumber \\
    &+ \lambda_{ws}^{tea}\mathcal{L}_{dice}(p_{tea}, y_p) + \lambda_{ws}^{tea^{\prime}}\mathcal{L}_{dice}(p_{tea}^{\prime}, y_p)
\end{align}

Building upon the above supervision, we further adopt two levels of consistency losses featuring the alignment of inter-model features with image embedding consistency loss and predictions under strong augmentation with the prediction consistency loss. 

The image embedding consistency loss $\mathcal{L}_{c}^{emb}$ ensures the correlation between the image embeddings generated by the student image encoder $z_{stu} = f(x; \mathit{\Delta^{stu}})$ and the teacher image encoder $z_{tea} = f(x; \mathit{\Delta^{tea}})$ with the same input. To achieve this goal, we adapt the widely-used InfoNCE method \cite{he2020momentum}, which has been utilized by many existing approaches \cite{zhao2023rcps, zhang2024improving} for contrastive learning that attracts positive pairs while simultaneously repelling negative pairs. In our work, we solely focus on encouraging the similarity between the image embeddings from two networks, which best suits our task. With an image embedding $z$, we extract the positive embeddings $z^{+}$  with the pseudo-label $y_p$:

\begin{equation}
z^{+}_i=\frac{\sum_{h, w} (y_p)_{h w} \cdot z_{i h w}}{\sum_{h, w} (y_p)_{j h w}}
\end{equation}

The image embedding consistency loss $\mathcal{L}_{c}^{emb}$ is then calculated by normalizing positive embeddings, which encourages similarity between the inter-model output positive pairs, the equation is formulated below with $\tau$ being the temperature coefficient:

\begin{equation}
\mathcal{L}_{c}^{emb}=-\log \sum_i \frac{ \exp \left(z^{stu+}_i \cdot z^{tea+} / \tau\right)}{\sum_i \exp \left(z^{stu+} \cdot z^{tea+} / \tau\right)}
\end{equation}

In addition, we employ the prediction consistency loss $\mathcal{L}_{c}^{pred}$ that imposes the prediction invariant of the teacher model when facing perturbed inputs. $\mathcal{L}_{c}^{pred}$ perform focal loss \cite{lin2017focal} and dice loss between $p_{tea}$ and $p_{tea}^{\prime}$ as an additional supervision of $\mathcal{L}_{sw}$:

\begin{equation}
\begin{aligned}
\mathcal{L}_{focal}(x,\hat{y}) & =-\sum_{i=1 \cdots N_T} \frac{1}{H W} \sum_{h, w} \mathbbm{1}\left(\hat{y}_{i h w}=1\right) \cdot \\
& \quad \quad \quad \quad \quad \quad \quad \quad \left(1-x_{i h w}\right)^\lambda \log \left(x_{i h w}\right) \\
& +\mathbbm{1}\left(\hat{y}_{i h w}=0\right) \cdot x_{i h w}{ }^\lambda \log \left(1-x_{i h w}^s\right)
\end{aligned}
\end{equation}

\begin{equation}
\mathcal{L}_{c}^{pred} = \lambda_{c}^{focal}\mathcal{L}_{focal}(p_{tea}^{\prime}, p_{tea}) + \lambda_{c}^{dice} \mathcal{L}_{dice}(p_{tea}^{\prime}, p_{tea})
\end{equation}

The final loss $\mathcal{L}_{total}$ for weak-supervised self-training is a linear combination of these four losses:

\begin{equation}
\mathcal{L}_{total} = \alpha\mathcal{L}_{ts} + \beta\mathcal{L}_{ws} + \gamma\mathcal{L}_{c}^{emb} + \delta\mathcal{L}_{c}^{pred} 
\end{equation}

Where $\alpha$, $\beta$, $\gamma$, and $\delta$ are weights controlling the balances between each loss.


\section{Experiment}

In this section, we begin by outlining the experimental settings and then present detailed comparisons with state-of-the-art methods and quantitative results. Finally, we analyze the effectiveness of individual components and specific network designs.

\subsection{Datasets}

We evaluate our proposed method on two X-ray fluoroscopy datasets: the publicly available Cardiac dataset \cite{gherardini2020catheter}, and the in-house Neurovascular dataset. The Cardiac dataset is a publicly accessible X-ray dataset comprising 2000 images derived from four fluoroscopy videos. These videos were recorded in an angiography suite during catheterization experiments conducted on a silicone aorta phantom. A semiautomated tracking method \cite{chang2016robust} was utilized to obtain annotations in the form of 2D coordinates of the catheter, confined to a manually selected region of interest. Our in-house Neurovascular dataset is obtained from experiments performed in a silicone anatomical phantom mimicking the intracranial vessels including the branches of the middle and anterior cerebral arteries at Geneva University Hospitals (Hôpitaux Universitaires de Genève). The images were acquired with an angiographic bi-plane C-arm (Azurion Philips, Best, the Netherlands). In order to have a constant opacification of the phantom arteries, we mixed in the reservoir of the circulating loop 25\% of contrast agent (Iopamiro 300 (Bracco, Milan, Italy)). This dataset contains scans with two orthogonal views with a region of 69.0 $mm$ $\times$ 69.0 $mm$. 

We divide each dataset into non-overlapping training and testing sets. In the coarse stage, even though the simulation data is unlimited, we limit the number of synthesized images to 254 to avoid prolonged training times and reduce the risk of overfitting. As for the fine stage, we adopt 112 training samples for the Cardiac dataset and 200 training samples for the Neurovascular dataset. Our method can achieve strong performance even with a limited number of training samples, resulting in reduced training costs and hours.

\begin{table*}[!htbp]
\centering
\begin{threeparttable}
\caption{Performance Comparison with State-of-the-Art Methods on the Cardiac Dataset. $^{*}$ Means Modifications Made to the Original Method Due to Poor Performance or Failure to Converge on Our Data. For Fully Supervised Methods, We Train the Model Using the Same Number of Samples as the UDA Methods.}
\label{sota_phantom}
\begin{tabular}{lcccccccc}
\hline 
\noalign{\smallskip}
\noalign{\smallskip}
Method & Backbone & Prompt Type & $\textit{IoU} \uparrow$ &  $F_1$  &  $Acc$ & $Sen$ \\
\hline 
\noalign{\smallskip}
\textbf{\textit{Fully-Supervised}} & & & & & & \\
\cline { 1 - 1 }
\noalign{\smallskip}
SAM (Real-label Supervised) \cite{kirillov2023segment} & Vit-B & end2end & 67.58 & 80.53 & 99.55 &81.21 \\
SAM (Pseudo-label Supervised) \cite{kirillov2023segment} & Vit-B & end2end & 61.32 & 75.71 & 99.44 & 77.57 \\
\noalign{\smallskip}
\hline
\noalign{\smallskip}
\textbf{\textit{Sim-to-Real Adaption}} & & & & & & \\
\cline { 1 - 1 }
\noalign{\smallskip}
Direct Transfer  & Vit-B & end2end & 53.84 & 69.99 & 99.29 & 71.46 \\
SAM \cite{kirillov2023segment} & Vit-B & box & 12.91 & 18.19 & 90.87 & 34.39 \\
SAM \cite{kirillov2023segment} & Vit-H & box &  18.82 & 26.29 & 90.44 & 68.91  \\
TENT \cite{wang2020tent} & Vit-B & end2end & 53.33 & 69.56 & 99.32 & 68.04 \\
TRIBE \cite{su2024towards}  & Vit-B & end2end & 54.02 & 70.04 & 99.30 & 71.77 \\
WeSAM$^{*}$ \cite{zhang2024improving}  & Vit-B & end2end  & 62.99 & 77.10 & 99.46 & 79.06 \\
\noalign{\smallskip}
\hline 
\noalign{\smallskip}
\textbf{Ours}  & Vit-B & end2end & \textbf{65.97} & \textbf{79.35} & \textbf{99.52} & \textbf{80.72} \\
\noalign{\smallskip}
\hline
\noalign{\smallskip}
\end{tabular}
\end{threeparttable} 
\end{table*}

\begin{table*}[!htbp]
\centering
\begin{threeparttable}
\caption{Performance Comparison with State-of-the-Art Methods on the Neurovascular dataset. $^{*}$ Means Modifications Made to the Original Method Due to Poor Performance or Failure to Converge on Our Data. For Fully Supervised Methods, We Train the Model Using the Same Number of Samples as the UDA Methods.}
\label{sota_hug}
\begin{tabular}{lcccccccc}
\hline 
\noalign{\smallskip}
\noalign{\smallskip}
Method & Backbone & Prompt Type & $\textit{IoU} \uparrow$ &  $F_1$  &  $Acc$ & $Sen$ \\
\hline 
\noalign{\smallskip}
\textbf{\textit{Fully-Supervised}} & & & & & & \\
\cline { 1 - 1 }
\noalign{\smallskip}
SAM (Real-label Supervised) \cite{kirillov2023segment} & Vit-B & end2end & 88.20 & 93.69 & 99.92 & 92.63 \\
SAM (Pseudo-label Supervised) \cite{kirillov2023segment} & Vit-B & end2end & 75.62 & 85.97 & 99.80 & 93.41 \\
\noalign{\smallskip}
\hline
\noalign{\smallskip}
\textbf{\textit{Sim-to-Real Adaption}} & & & & & & \\
\cline { 1 - 1 }
\noalign{\smallskip}
Direct Transfer  & Vit-B & end2end & 70.81 & 82.90 & 99.73 & 91.87\\
SAM \cite{kirillov2023segment} & Vit-B & box & 18.15 & 23.19 & 92.54 & 40.60 \\
SAM \cite{kirillov2023segment} & Vit-H & box &  25.09 & 32.11 & 92.33 & 75.75  \\
TENT \cite{wang2020tent}  & Vit-B & end2end & 69.93 & 81.89 & 99.71 & 93.38 \\
TRIBE \cite{su2024towards}  & Vit-B & end2end & 70.76 & 82.87 & 99.72 & \textbf{91.96} \\
WeSAM$^{*}$ \cite{zhang2024improving}  & Vit-B & end2end  & 76.84 & 86.72 & 99.82 & 90.61 \\
\noalign{\smallskip}
\hline 
\noalign{\smallskip}
\textbf{Ours}  & Vit-B & end2end & \textbf{81.92} & \textbf{89.99} & \textbf{99.86} & 91.44 \\
\noalign{\smallskip}
\hline
\noalign{\smallskip}
\end{tabular}
\end{threeparttable} 
\end{table*}

\subsection{Experiment Details}

We chose Vit-B \cite{dosovitskiy2020image} as the encoder network for all the SAM models in this work due to memory constraints and the need for real-time application. The prompt in the training phrase is generated using the pseudo-labels from the coarse stage. Specifically, we use the minimal bounding box that covers the entire instance segmentation mask as the box prompt. As an alternative, point prompt is generated by randomly selecting $n$ positive points within the instance segmentation mask and $n$ negative points outside. For test and inference, the student model performs in an end-to-end fashion that outputs the segmentation mask without the need for human interactions. We report the widely adopted Intersection Over Union (\textit{IoU}), $F_1$ score ($F_1$), accuracy ($Acc$), and sensitive score ($Sen$) as the evaluation metrics of our method.

For the hyperparameters, we adopt the Adam optimizer with the learning rate set to 0.0001. The batch size is set to 2 for training and 16 for testing with one node of 80GB RTX A100 GPU. The loss weights $\lambda$s in $\mathcal{L}_{c}^{pred}$ and $\mathcal{L}_{ws}$ are all set to 0.5, whereas in $\mathcal{L}_{ts}$, they are set to 1. $\lambda$ in $\mathcal{L}_{focal}$ and $\tau$ in $\mathcal{L}_{c}^{emb}$ are set to 2 and 0.3, respectively. To balance out each loss, $\alpha$, $\beta$, $\gamma$, and $\delta$ are set to 0, 1, 0.5, and 1, respectively. After the warm-up epochs of 3, the weights are adjusted to $\alpha = 5$, $\beta = 0$, $\gamma = 1$, and $\delta = 0$. For strong data perturbation $\mathcal{P}$, we follow the same pipeline as in \cite{xu2021end}.

\subsection{Quantitative Evaluations}

We primarily present the quantitative results for the Cardiac dataset and the Neurovascular dataset, comparing them with the current state-of-the-art outcomes shown in Table \ref{sota_phantom} and Table \ref{sota_hug}, respectively. We evaluate multiple domain adaptation approaches in our sim-to-real settings using the exact same training setups as adopted by our method for fair comparison. To be more specific, we compare our method with Direct Transfer, out-of-the-box SAM \cite{kirillov2023segment} with box prompt, TENT \cite{wang2020tent}, TRIBE \cite{su2024towards}, and WeSAM \cite{zhang2024improving}. Direct Transfer means we test the source domain pre-trained model on the target domain directly without any adaption. To evaluate TENT and TRIBE, we adopt their Test-Time Adaptation (TTA) technique on top of a SAM fine-tuned on the source domain. WeSAM is a recent self-training-based strategy to adapt SAM to target distribution, but directly utilizing their implementation on our data leads to poor performance and convergence issues. We adapted their method by substituting their anchor SAM with the supervision of pseudo-labels, and we trained their framework in an end-to-end manner, eliminating the need for human interaction during testing as described in their paper.

As demonstrated in Table \ref{sota_phantom}, our proposed method achieves a significant performance enhancement over existing Unsupervised Domain Adaptation (UDA) techniques on the Cardiac dataset, outperforming the next best approach by a notable margin of 2.98\% in \textit{IoU} and is very close to the fully-supervised upper bound of 67.58\%. Furthermore, our method can dramatically improve the Direct Transfer method by 12.13\% in \textit{IoU}, which translates to an overall increase of 22.53 percentage points. It is also noteworthy to highlight that our approach exceeds the performance of the pseudo-label supervised SAM by 4.65\%. This result illustrates the superior quality of supervision provided by our teacher model in comparison to pseudo-labels, validating the robustness and reliability of our method.


The evaluation results on the Neurovascular dataset, presented in Table \ref{sota_hug}, also demonstrate that our method achieves the highest performance among state-of-the-art UDA techniques. Specifically, our approach surpasses the Direct Transfer baseline by a significant margin of 6.3\%. When compared to other advanced methods, such as the modified WeSAM, TENT, and TRIBE, our framework outperforms them by substantial margins of 5.08\%, 11.99\%, and 11.16\% measured in \textit{IoU}, respectively. Although there remains a gap between our method and the fully supervised approach, it is noteworthy that our framework still exceeds the pseudo-label supervised SAM by 6.3\%. This notable improvement is attributed to the superior quality of supervision provided by our teacher SAM model, which ensures more accurate and reliable results.

\begin{figure}[t]
  \centering
  \includegraphics[width= \linewidth]{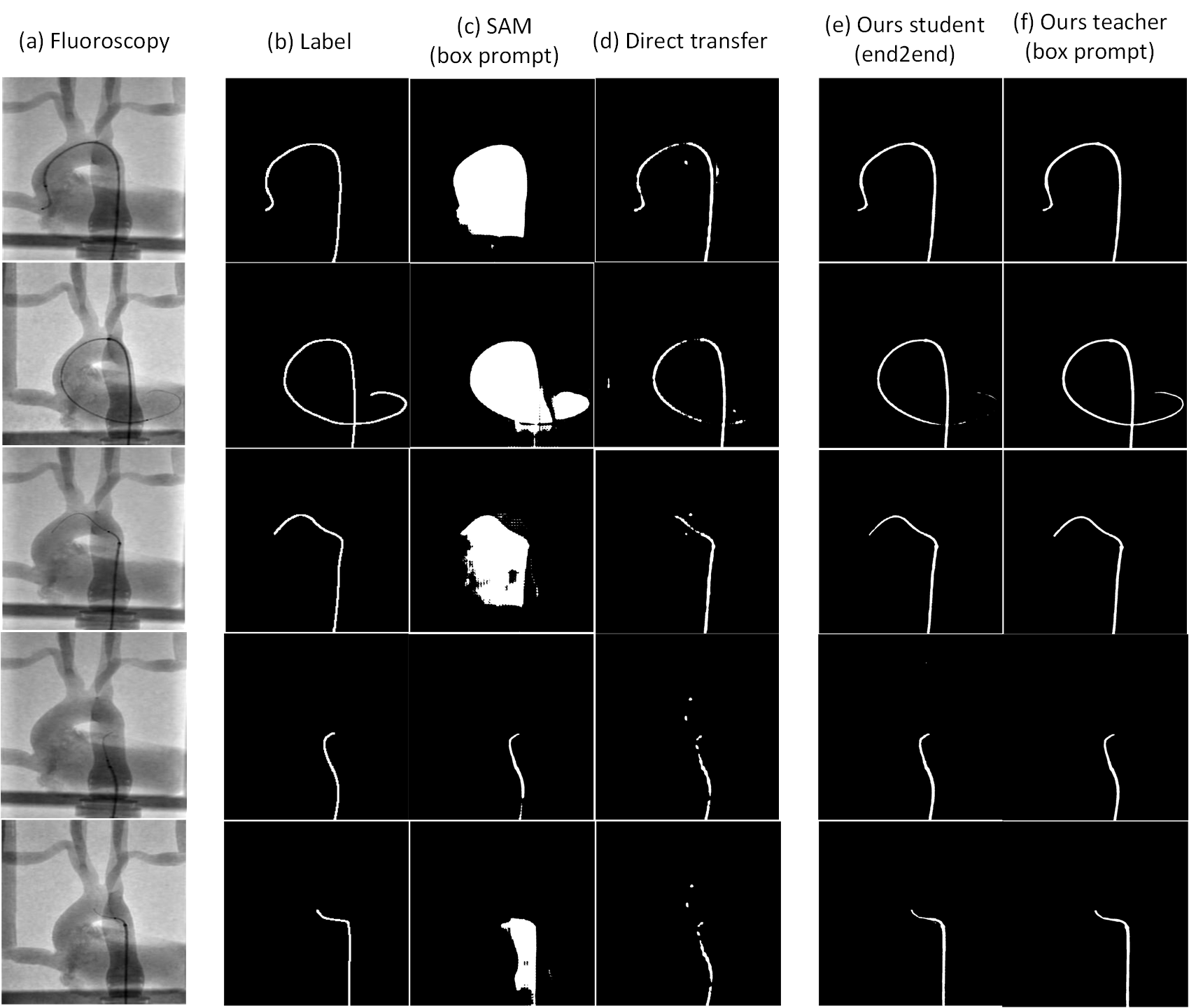}
  \caption{Visual comparison on the Cardiac dataset. (a) is the original X-ray fluoroscopy image, (b) is the corresponding label, (c), (d), (e), and (f) are the prediction results from SAM \cite{kirillov2023segment} with box prompt, Direct Transfer, our end-to-end student model, and our teacher model with box prompt, respectively.}
  \label{vis_phanton}
\end{figure}

\begin{figure}[t]
  \centering
  \includegraphics[width= \linewidth]{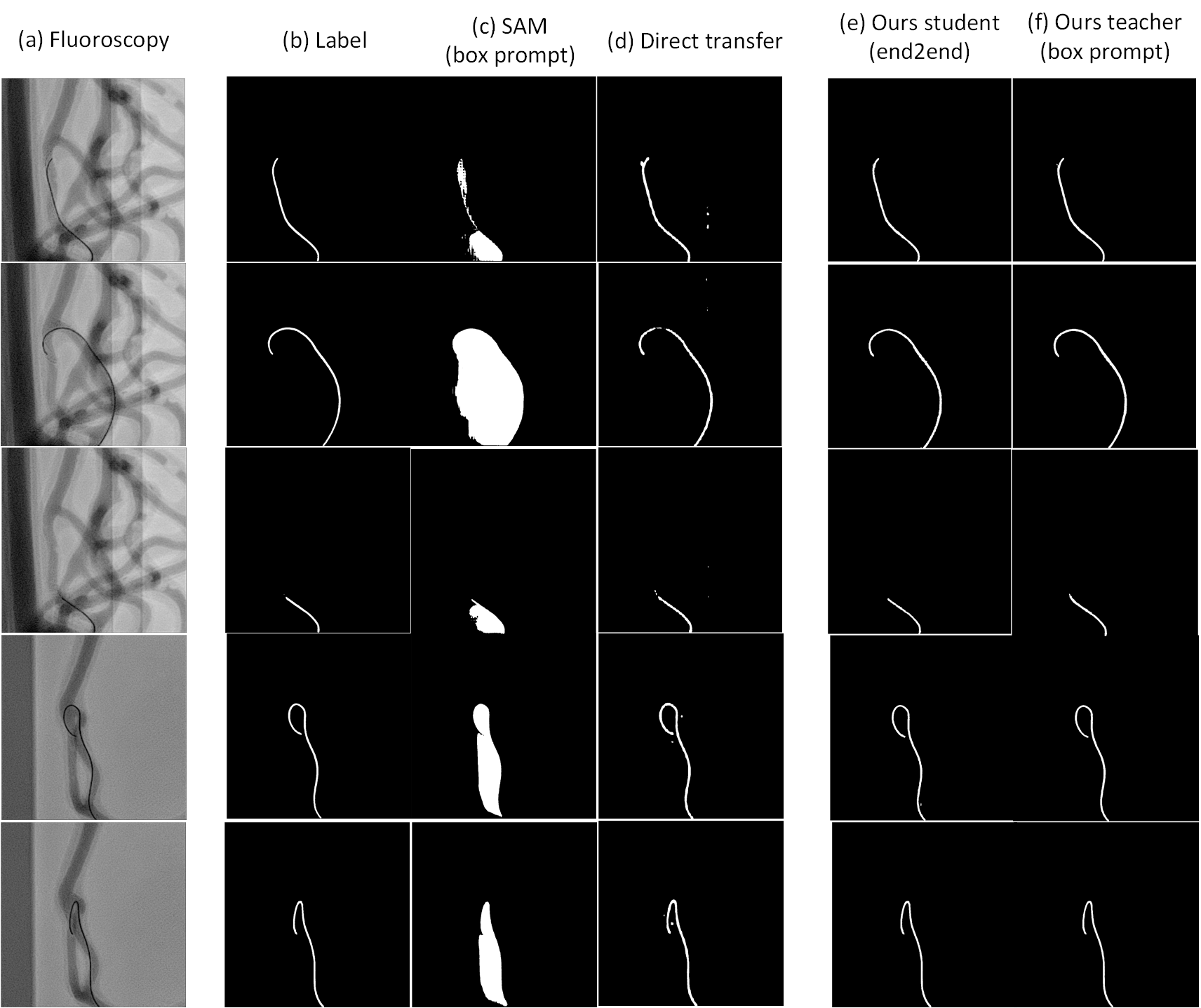}
  \caption{Visual comparison on the Neurovascular dataset. (a) is the original X-ray fluoroscopy image, (b) is the corresponding label, (c), (d), (e), and (f) are the prediction results from SAM \cite{kirillov2023segment} with box prompt, Direct Transfer, our end-to-end student model, and our teacher model with box prompt, respectively.}
  \label{vis_HUG}
\end{figure}

\subsection{Visualization Evaluations}

In addition to the quantitative evaluations, we further conduct visualization evaluations to intuitively illustrate the performance of the proposed method. In Fig. \ref{vis_phanton} and Fig. \ref{vis_HUG}, we show the prediction results from a box-prompted SAM \cite{kirillov2023segment} and Direct Transfer in addition to our end-to-end student model and our box-prompted teacher model for comparison in both the Cardiac dataset and the Neurovascular dataset. Due to the thin and long shape of the guidewire, the out-of-the-box SAM performs awkwardly even with prompt guidance, outputting a large chunk of masks that completely miss the target. The Direct Transfer method cannot preserve guidewires’ smoothness and continuity, whereas our method can generate smooth and intact segmentation masks in various complex cases like large curvature, occlusion of the vessel, and deformed tips. Additionally, our teacher model can generate more intact predictions with box prompts, successfully capturing the thinnest guidewire tip in challenging cases.

\subsection{Ablation Study}

In this section, we thoroughly analyze the effectiveness of each individual component within our framework on the Cardiac dataset, the results are shown in Table \ref{aba}. When the coarse stage is enabled to provide weak supervision $\mathcal{L}_{ws}$, a significant performance gain is observed (12.91\% $\rightarrow$ 62.71\%) over the SAM baseline with box prompting, highlighting the significance of directing the model's learning process. In addition, incorporating the student-teacher network along with the teacher model's supervision $\mathcal{L}_{ts}$ on top of the coarse-to-fine pipeline brings an additional 2.88\% improvement with its further supervision. This improvement highlights the effectiveness of the teacher model's more refined supervision compared to the pseudo-labels. Finally, the two consistency losses, $\mathcal{L}_{ws}$ and $\mathcal{L}_{c}^{emb}$, further aid the learning process, each contributes its own improvements, underscoring their necessity. These results collectively emphasize the necessity of each component in optimizing the overall performance of our method.

\begin{table}[h]
\centering
\caption{Component Analysis of the Proposed Framework.}
\label{ca}
\begin{tabular}{cccccccc}
\toprule
\textit{Stu-Tea} & $\mathcal{L}_{ws}$ & $\mathcal{L}_{c}^{emb}$ & $\mathcal{L}_{c}^{pred}$ & 
$\textit{IoU} \uparrow$ & 
$F_1$ & $Acc$ &
$Sen$ \\
\midrule 
\noalign{\smallskip}
\multicolumn{4}{c}{\textit{SAM (Vit-b)}} & 12.91 & 18.19 & 90.87 & 34.39\\ 
$\sqrt{}$ & & & & 21.75 & 29.16 & 91.91 & 55.57 \\
 & $\sqrt{}$ & & & 62.71 & 76.91 & 99.44 & 81.62 \\
$\sqrt{}$ & $\sqrt{}$ & & & 65.59 & 79.82 & \textbf{99.52} & \textbf{82.89} \\
$\sqrt{}$ & $\sqrt{}$ & $\sqrt{}$ & & 65.69 & 79.16 & 99.50 & 82.81 \\
$\sqrt{}$ & $\sqrt{}$ &  & $\sqrt{}$ & 65.73 & 79.19 & 99.50 & 82.64 \\
$\sqrt{}$ & $\sqrt{}$ & $\sqrt{}$ & $\sqrt{}$ &  \textbf{65.97} & \textbf{79.35} & \textbf{99.52} & 80.72  \\
\noalign{\smallskip}
\bottomrule
\noalign{\smallskip}
\label{aba}
\end{tabular}
\\
\end{table}

\subsection{Additional Analysis}

\subsubsection{Visualization of the Training Strategy and Learning Process}

\begin{figure}[t]
  \centering
  \includegraphics[width=0.95 \linewidth]{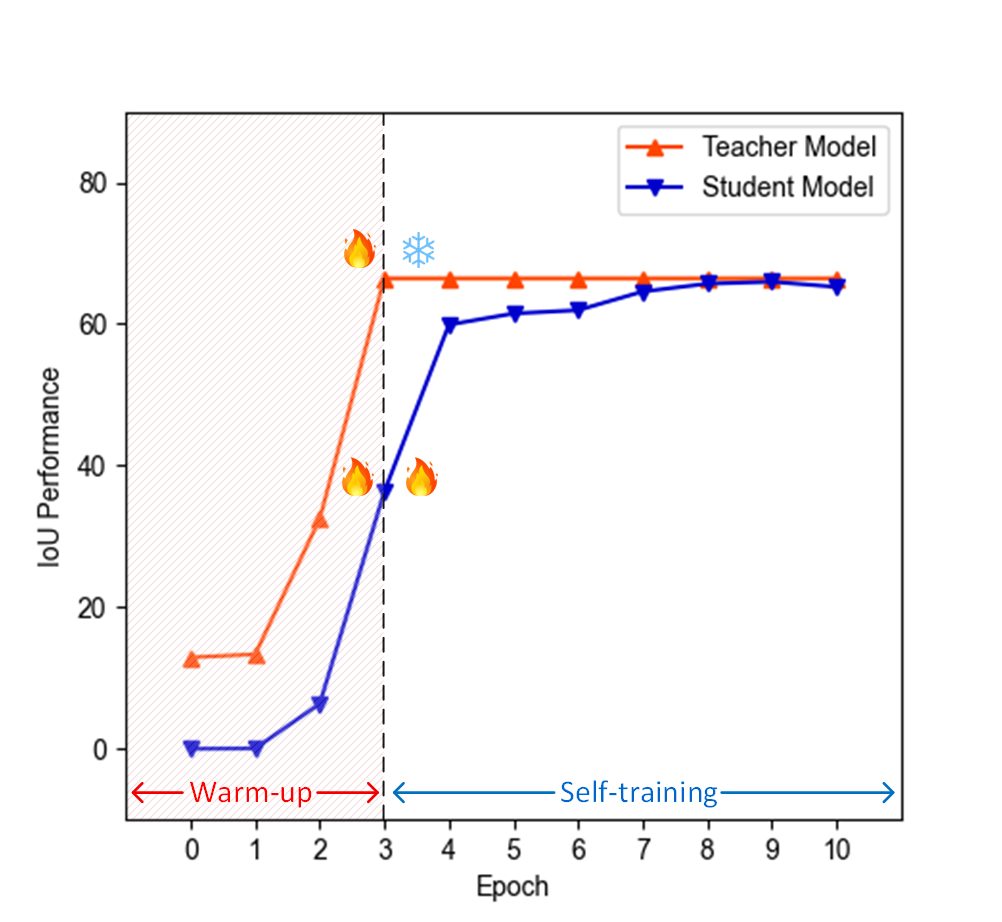}
  \caption{Visualization of the learning process and the self-training strategy. The evolution of the \textit{IoU} scores highlights the convergence of both the teacher and student network in the fine stage. The ``fire'' symbol means the trainable parameters, while the  ``snowflake'' symbol means the parameters are frozen without updating. We froze the LoRA parameters in the image encoder of the teacher model after warmup to prevent the negative impact of self-training.}
  \label{figepoch}
\end{figure}

To provide a clearer understanding of our training strategy for the teacher-student network during the fine stage, we have plotted the evolution of the Intersection over Union (\textit{IoU}) performance across training epochs, as shown in Fig. \ref{figepoch}. The graph illustrates our two-phase training approach: an initial warm-up phase followed by a self-training phase. During the warm-up phase, both the teacher and student networks are trained using pseudo-labels. During this stage, the teacher network can quickly converge with the guidance from box prompts. Then the teacher model's predictions are leveraged to supervise the student network in the subsequent self-training phase. This strategy enables the student network to progressively refine its performance, ultimately achieving a level of performance comparable to that of the teacher model. The resulting performance clearly demonstrates the effectiveness of our phased approach in optimizing the outcomes of both networks.

\subsubsection{Impact of the Number of Target Domain Training Samples Used on Fine Stage}

\begin{figure}[t]
  \centering
  \includegraphics[width=0.95 \linewidth]{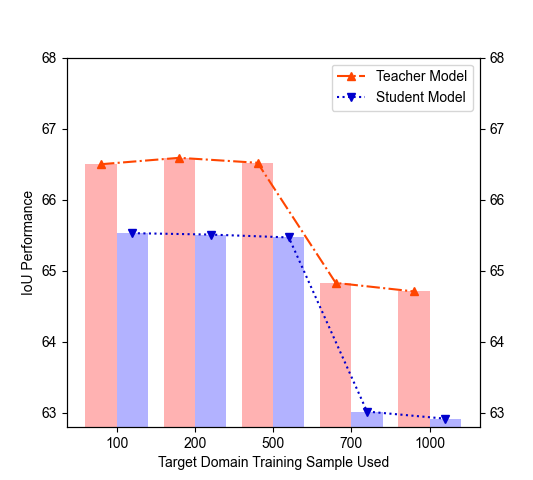}
  \caption{Impact of target domain training samples size on the performance of both the teacher model with box prompting and the end-to-end student model. }
  \label{fig_sample}
\end{figure}

After generating pseudo-labels through direct transfer in the coarse stage, the number of samples used to fine-tune the student-teacher network becomes a crucial factor influencing overall performance. Utilizing too few samples can hinder the network’s ability to converge, while using an excessive number of samples can introduce risks of overfitting or degraded results, particularly if the pseudo-labels are of poor quality or contain misleading information. This trend is illustrated in Fig. \ref{fig_sample}, where we observe that employing 100 (5\%) to 500 (25\%) samples achieves an optimal balance between performance and training efficiency, yielding the best results. However, when the training set is expanded to include 35\% of the total samples, the performance of the teacher network declines by approximately 1.5\% in \textit{IoU}, which subsequently leads to a similar decrease in the student model’s performance.

\subsubsection{Inference With Different Prompt Types}
\label{IV.F.3}

In addition to the student model’s end-to-end prediction capabilities, it also demonstrates versatility by effectively generalizing to take in different types of prompts as guidance, such as box and point prompts. We conducted an evaluation to compare the performance of both the student and teacher models using various prompt types: box prompts, point prompts with 5 positive and 5 negative points, and a combination of both. The results, presented in Table \ref{prompt}, show that the student model achieves better performance when using both point and box prompts, attaining a 66.32\% \textit{IoU}. This result is notably close to the teacher model’s upper bound of 66.50\% \textit{IoU}, with a minimal difference of just 0.18\%. This finding highlights the student model’s ability to closely approximate the teacher model's performance when provided with additional guidance through prompts.

\begin{table}[h]
    \centering
    \caption{Performance Evaluation of the Teacher and Student models With Different Prompt Types}
\begin{tabular}{lcccc}
\toprule  Prompt Type & $\textit{IoU} \uparrow$ & $F_1$ & $Acc$ & $Sen$ \\
\midrule \textit{SAM} & & & & \\
\cline { 1 - 1 } \noalign{\smallskip} Box & $\mathbf{12.91}$ & 18.19 & 90.87 & 34.39 \\
Point & 9.03 & 15.73 & 90.70 & 54.22 \\
Point + Box & 11.22 & $\mathbf{19.20}$ & $\mathbf{92.43}$ & $\mathbf{60.75}$ \\
\midrule \textit{Teacher Model} & & & & \\
\cline { 1 - 1 } \noalign{\smallskip} Box & $\mathbf{6 6 . 5 0}$ & $\mathbf{7 9 . 7 4}$ & $\mathbf{9 9 . 5 1}$ & 82.66 \\
Point & 64.66 & 78.21 & 99.45 & $\mathbf{8 4 . 7 7}$ \\
Point + Box & 65.53 & 79.02 & 99.48 & 84.73 \\
\midrule \textit{Student Model}& & & & \\
\cline { 1 - 1 } \noalign{\smallskip} End-to-end & 65.97 & 79.35 & $\mathbf{99.52}$ & 80.72 \\
Box & 65.53 & 79.03 & 99.51 & 79.86 \\
Point & 65.91 & 79.32 & 99.51 & 80.88 \\
Point + Box & $\mathbf{6 6 . 3 2}$ & $\mathbf{7 9 . 6 1}$ & $\mathbf{9 9 . 5 2}$ & $\mathbf{81.03}$ \\
\bottomrule
\end{tabular}
\label{prompt}
\end{table}

\subsubsection{Hyper-Parameter Selection}


In Table \ref{tabHype}, we analyze the sensitivity of our model to various hyperparameters, specifically focusing on the learning rate ($Lr$) and the temperature coefficient ($\tau$) used in the embedding consistency loss. For the learning rate, we employed the Adam optimizer with values set at 1e-3, 1e-4, and 1e-5. For the temperature coefficient $\tau$, we compared a range of values, including 0.1, 0.3, 0.5, 0.7, and 0.9. Our results indicate that the proposed sim-to-real adaptation method exhibits a relatively stable performance across different hyperparameter settings. The model consistently achieves strong results with learning rates between 1e-3 and 1e-4, and with temperature coefficients ranging from 0.3 to 0.9. This robustness suggests that our method is not overly sensitive to hyperparameter variations, making it more adaptable and easier to fine-tune for different datasets or tasks.

\begin{table}[h]
    \centering
    \caption{Study on Hyper-Parameter Selection of the Learning Rate ($Lr$) And the Temperature Coefficient ($\tau$).}
\begin{tabular}{cc|cccc}
\toprule \multicolumn{2}{c|}{Hyper-Parameters} & $\textit{IoU} \uparrow$ & $F_1$ & $Acc$ & $Sen$ \\
\midrule \multirow{3}{*}{\textit{Lr}} & $1 \mathrm{e}-3$ & 65.01 & 78.66 & 99.49 & 81.46 \\
& $\mathbf{1 e}-4$ & $\mathbf{6 5 . 9 4}$ & 79.35 & 99.51 & 82.66 \\
& $1 \mathrm{e}-5$ & 62.56 & 76.73 & 99.45 & 80.96 \\
\midrule \multirow{5}{*}{$\tau$} & 0.1 & 64.87 & 78.53 & 99.49 & 81.06 \\
& $\mathbf{0 . 3}$ & $\mathbf{6 5 . 9 4}$ & 79.35 & 99.51 & 82.66 \\
& 0.5 & 65.42 & 78.90 & 99.51 & 79.42 \\
& 0.7 & 65.45 & 78.92 & 99.51 & 79.57 \\
& 0.9 & 65.41 & 78.89 & 99.51 & 79.52 \\
\bottomrule
\end{tabular}
\label{tabHype}
\end{table}

\section{Discussion and Future Work}

Our work demonstrates the potential to generalize a vision foundation model for real-world X-ray guidewire segmentation by exclusively relying on simulation data, eliminating the need for any annotation. Despite the success of our sim-to-real framework, there exist possible limitations that need to be tackled in future research. To begin with, the relationships between frames are overlooked. While this leads to more lenient requirements for the input data, modeling frame-to-frame interactions and taking video streams instead of X-ray images as inputs could enhance performance and improve continuity. Additionally, a performance decline is observed when increasing the number of training samples during fine-tuning, likely due to overfitting or training hyperparameters that are not universally effective across larger datasets.

To address these limitations, we propose the following future improvements: 1) Develop a new model version capable of processing video streams as input. 2) Investigate the cause of the performance decline with more training samples by testing alternative hyperparameters or diversifying the dataset through additional augmentations.

\section{Conclusion}


In this paper, we introduce a novel sim-to-real coarse-to-fine domain adaptation framework designed to generalize an end-to-end vision foundation model for application in real-world X-ray fluoroscopy, eliminating the need for manual annotations in the target domain. In the coarse stage, pseudo-labels are generated through transfer learning by utilizing a source image style transfer technique that maintains the integrity of the guidewire structure. Building upon this foundation, we further leverage weak supervision and self-training within a teacher-student network to achieve results that surpass the coarse labels. We conduct extensive experiments on two real-world X-ray fluoroscopy datasets, demonstrating the effectiveness and superiority of our proposed method in this challenging domain.

\section*{Acknowledgments}

This work is supported by the internal funding of EPFL (École polytechnique fédérale de Lausanne) and the SNSF / Swiss National Science Foundation 320030$\_$188942.


\bibliographystyle{IEEEtran}
\bibliography{Ref}


 




\vfill

\end{document}